# Non-technical Loss Detection with Statistical Profile Images Based on Semi-supervised Learning

Jiangteng Li [1] and Fei Wang [2,*]

[1] School of Electronic Information and Communications, Huazhong University of Science and Technology, Wuhan 430074, China; lijiangteng@hust.edu.cn
[2] School of Electronic Information and Communications, Huazhong University of Science and Technology, Wuhan 430074, China; wangfei@hust.edu.cn
* Correspondence: wangfei@hust.edu.cn

**Abstract:** In order to keep track of the operational state of power grid, the world's largest sensor systems, smart grid, was built by deploying hundreds of millions of smart meters. Such system makes it possible to discover and make quick response to any hidden threat to the entire power grid. Non-technical losses (NTLs) have always been a major concern for its consequent security risks as well as immeasurable revenue loss. However, various causes of NTL may have different characteristics reflected in the data. Accurately capturing these anomalies faced with such large scale of collected data records is rather tricky as a result. In this paper, we proposed a new methodology of detecting abnormal electricity consumptions. We did a transformation of the collected time-series data which turns it into an image representation that could well reflect users' relatively long term consumption behaviors. Inspired by the excellent neural network architecture used for objective detection in computer vision domain, we designed our deep learning model that takes the transformed images as input and yields joint featured inferred from the multiple aspects the input provides. Considering the limited labeled samples, especially the abnormal ones, we used our model in a semi-supervised fashion that is brought out in recent years. The model is tested on samples which are verified by on-field inspections and our method showed significant improvement.

## 1. Introduction

By deploying a large number of sensor devices, the smart grid becomes the largest sensor network in world. Particularly, State Grid Corporation of China(SGCC) has installed nearly 500 million smart meters (SM) by the end of 2017. The smart meters have greatly improved the information and automation level of the power grid, and making it possible to identify anomaly. Anomalies, which include meters failure and electricity theft, are the primary source of non-technical losses (NTL) in power grid. The NTL would cause significant revenue losses, $58.7 billion dollars are lost each year due to NTL [26]. Further, it also affects the power system operation because of the uncertainty of the real consumption [1]. Due to the huge scale of the collected data and the complicated real-world operational environment, the NTL detecting methods are required that are excellent in terms of both efficiency and reliability. However, the existing expert system based approaches sometimes are unreliable as it fails to comprehensively exploit the underlying correlations among different features. Or the other way round, in order to cover extremely diverse forms of abnormal behaviors, far more field knowledge should be integrated, and sometimes human efforts are necessary. As a result, efficiency suffers.

In this paper, we introduced deep learning to help with the detection of NTL as it has already demonstrated its great potential and capacity in a variety of fields. Neural networks' hierarchical structures are designed to automatically extract higher level features containing more abstract semantic information from large scale of input data. So we believe in our problem setting where the prominent features are latent and intertwined with a lot of factors, deep neural networks may just be fit for the job in terms of both efficiency and accuracy. Yet, there are problems we are to solve :

a) The quality of collected data is relatively low because of data noise and data missing. As is often the case in real-world scenarios, the data records may suffer from various types of noises. Generally, two types of noises are distinguished: feature (or attribute) noise and class noise. Class noise means the labels assigned to samples are not totally correct, which is likely to happen as the ground truth labels are given based on the result of on-field inspections in this problem. In this paper we only focus on the feature noises. Generally speaking, this type of noise can be summarized as the noise that interferes the mapping from the input $x$ to the output label $y$, including slight value disturbance, erroneous data records as well as data out of sync. Data missing is also a common phenomenon in large-scale sensing systems. Situations get worse when unavailable data points emerge in high density or even large chunks are missing.

b) The correlation between labels given by on-field inspection and the original data records is not intuitional. For those verified customers, we only have general labels denoting whether anomalies ever occurred in a customers' history records. There is no further information that indicates when and how faults or fraudulent events occurred. So we have to find distinct patterns of positive/negative samples for the classifier/detector. It is better if we can make the data records more human friendly and distinguishable for ordinary people without the aid of carefully designed criteria.

c) The proportion of labeled the data is rather small compared to that of the unlabeled. And the abnormal samples are obviously fewer than normal ones. This fact makes it impossible for us to treat the problem as a simple supervised classification problem.

In this paper, we first proposed a data transformation method. By analyzing the abnormal cases, we found that statistical characteristics of the data can better reflect the consumption behaviors of the electricity users. The method we designed focuses on the feature patterns in a relatively longer term by utilizing the data in a certain time range. This transformation makes it easier to get features that can distinguish between normal and abnormal samples. Moreover, the transformed data form naturally has better noise resistance performance and it can eliminate the difficulties resulting from different value ranges of different users. We also adopt a semi-supervised model brought out in recent years. This model does not depend on any presupposed prior distribution of the training data in class space and proved itself a reliable one by testing it on the real customers' data.

There is an abundance of previous work about NTL detection or anomaly (fraud) detection in electricity industry. The approaches can generally be categorized in classification based and non-classification based solutions. The non-classification based approaches usually use methods like clustering, statistical analysis and so on. In [2,3], they forecast the customers' energy consumption. If the difference between actual and forecasted consumption exceeds the limit imposed by the authors, the customer is considered to be committing fraud. Another similar approach in [4,5], is to first construction the distribution of the historical data collected by meters. And then a similarity measurement is introduced to find outliers, which are considered fraud or anomaly. As far as the classification based approaches are concerned, Nagi et al. [6] used extracted features from historical customer consumption data and trained a SVM model with them to detect NTL. Apart from the sole consumption measurements, P. Glauner et al. [7] also utilized the customers' geographical location to compute the inspection rate and the NTL rate in its neighborhood. And other models besides SVM, such as LR, KNN, RF are trained to testify the methodology. As for the classifier, there also exists works used neural networks (NN) as in [8] and [9]. However, [9] did a comparison among different algorithms and found out the single gradient boosted machine (GBM) outperformed any ensemble or any other classifiers.

Considering the afore-mentioned approaches, our contributions to these problems are as follows:

- We designed a novel way of visualizing customers' data records utilizing more potentially useful real-time electrical information than sole energy consumption measurements. The new data format statistically constructs a visual representation of customers' behavior in a certain time range and provides a new aspect of viewing their operational states. Such transformation is also tolerant of slight data error or missing, which is quite common for the sensor systems in industrial areas. The convolutional based neural network that is universally used in computer vision domain is then introduced to extract visual patterns for further classification.
- The inefficient on-field inspection limited the ground truth we can make use of for supervised training. This work combined a semi-supervised learning method based on consistency loss with our proposed network architecture to deal with the common situations in real world scenarios where the training samples are partially labeled. It turns out the consequent overfitting problem can be addressed pretty well.

**2. Data Transformation and network design**

The major goal that our methodology is going to achieve in this paper is to decide whether there exists abnormal energy consumption for a certain user, caused by metering failure or fraudulent usage, by utilizing the SM records. And the data records, indexed by unique customers' identification numbers, are from an electrical information collection system constructed by a power company in a certain province of China.

For the customers we are dealing with, AC electric parameters are what we mainly focus on. Compared with the energy consumption that most previous work dealt with, AC parameters, including three phase voltage, current, power factor, active power and reactive power, have higher collection frequencies. We believe features with higher frequencies allow us to get detecting results with lower delay thus making quicker response to the NTLs. Feature engineering is the first step to expose most prominent information as well as bridge the discrepancies between customers such as differences in magnitudes and value ranges for a single attribute. Then we form our training samples based on extracted features and build a proper inference network to get an appropriate feature representation. And the network is trained in a semi-supervised fashion considering the fact that the labeled samples are in the minority.

**Table 1.** AC electric parameters

| Items | Notes |
|---|---|
| $U_A, U_B, U_C$ | Voltage on three phases |
| $I_A, I_B, I_C$ | Voltage on three phases |
| Active power | |
| Power factor | |

We randomly select a small portion of all the verified customers together with far more unverified customers. For validation, the rest of the verified customers are adopted to evaluate the performance of our method. The random customer selection and training process is executed for multiple times to make our results more reliable.

*2.1. Statistical profile*

The abnormal consumption that really matters and would cause NTL is the kind that last for at least a period of time. To be specific, anomalies that take on the forms of impulse signals, sudden changes or transient data missing for example, would not make any significant differences to the overall energy consumption (EC) of a customer, neither would them cause great non-technical losses. It is a better choice to take them as the consequences of communication failure or temporal metering devices' malfunctioning and treat the data records from a statistical point of view.

What are presented in Figure 1 are the three-phase voltage (normalized) records from two customers, A and B. And by on-field inspections they are considered a normal and an abnormal customer respectively. From the time series in (a) we can see that for the normal customer A (upper), apart from several sudden drops in the single-phase voltage, the voltage values on three phases are highly identical for the rest of the time. However, for customer B, it is a consistent and almost periodic phenomena that voltage on phase c differs significantly from the rest of the phases. From the scatter plot of customer A in (b), the several outliers correspond to those impulses in (a) and they behave more like noises in this case. This justifies what was discussed above: focusing on the long term consumption behaviors would help us draw saver conclusions in this NTL detection problem. So assigning the customers' labels to each row of data records and taking it as an input/label training sample pair would introduce a lot of noises that deteriorate the training. And this is why we proposed our data transformation method. Further experimental comparison between single-collecting-point sample and our statistics-based data representation would be shown in later section.

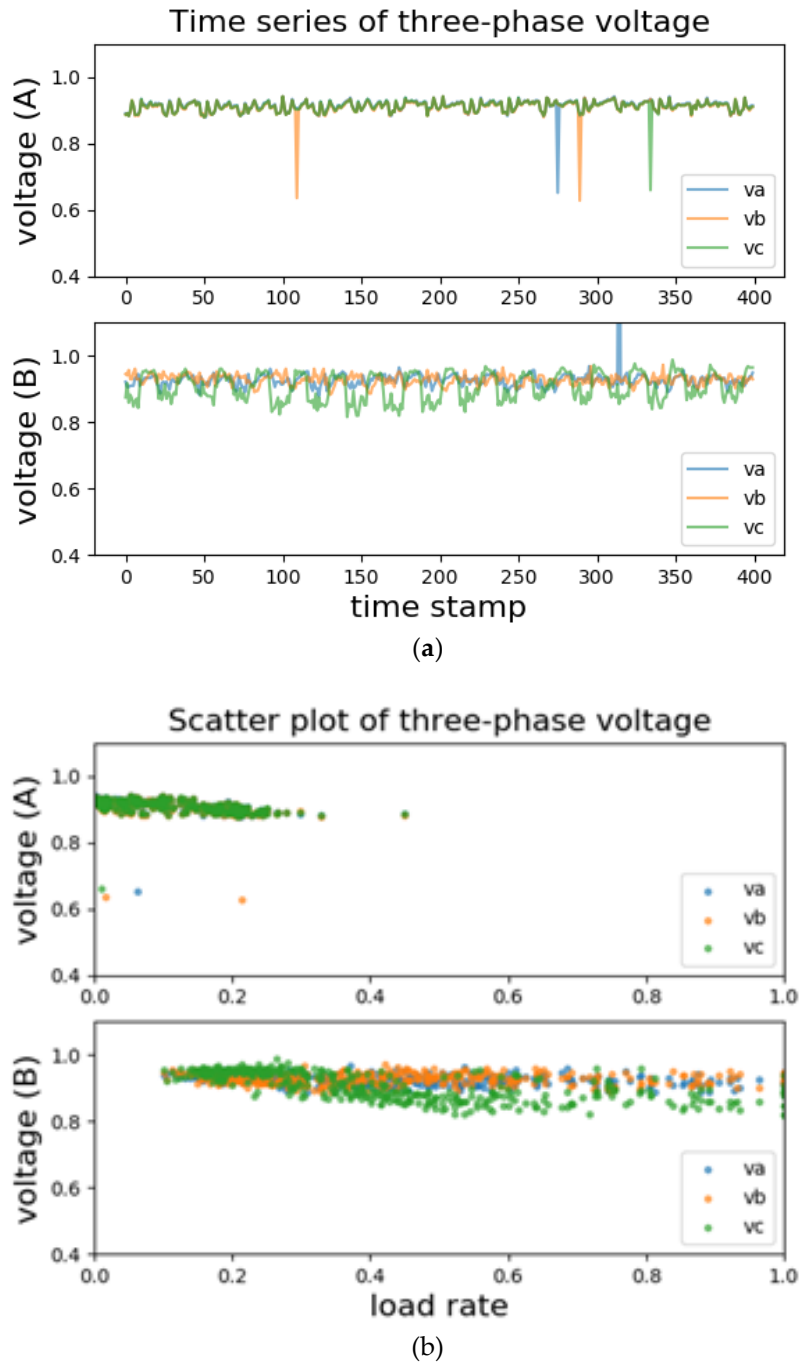

**Figure 1.** (a) Two time series from a normal (upper) and an abnormal (lower) customer respectively; (b) The corresponding scatter plots of the same piece of data with load rate being the x-axis.

Here, we did not aim to get a precise mathematical representation of the feature distributions. Rather, we designed a data transformation inspired by kernel density estimation (KDE), a non-parameter distribution estimation method. In this work, we did a 2-dimensional KDE transformation of the features obtained in previous section.

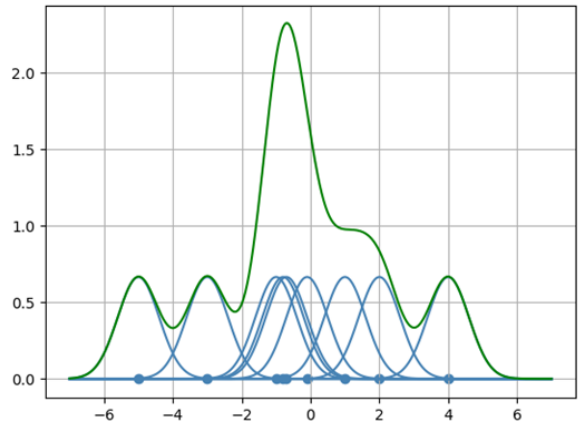

(a)

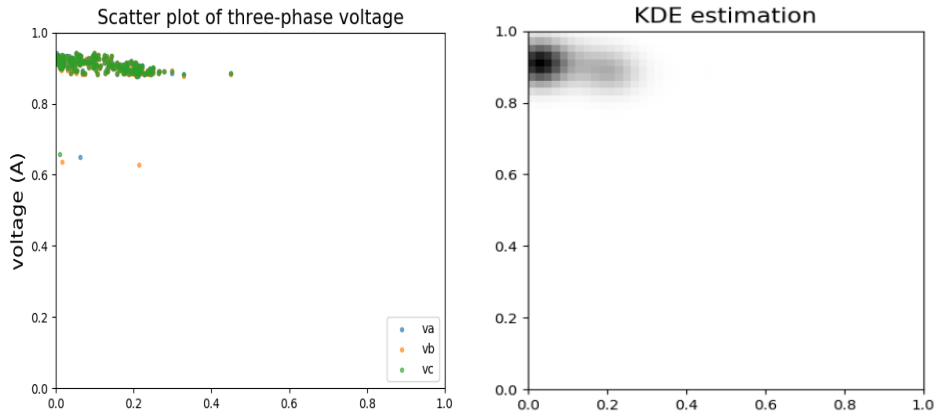

(b)

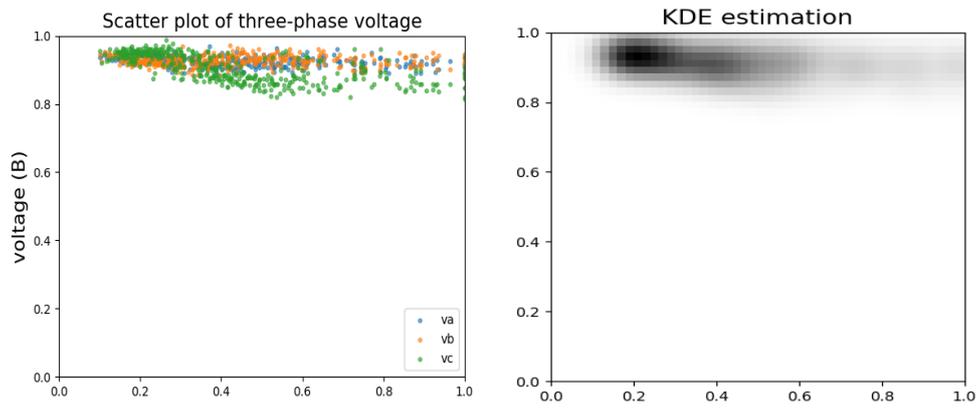

(c)

**Figure 2.** (a) An example of 1-D KDE with Gaussian kernels. The blue points are the actual data points collected. And they are treated as the mean values of the Gaussian curves in blue. The green curve is the final estimation of data distribution by summing up all the kernel curves. (b) Scatter plot (left) and statistical profile using KDE (right) of the normal customer A. (c) Scatter plot (left) and statistical profile using KDE (right) of the normal customer B.

The process can be briefly thought of as replacing each discrete point in a scatterplot with a kernel distribution. The idea of this estimation method is that the positions where the points appear have the higher occurrence probabilities. So, for the Gaussian distribution we use as basic kernels, the peaks are where the points are located. The advantage of distribution profile over scatter plot is that it gives more intuitive impression of customers' behavior patterns during a period of time. With the extension of time used for statistical analysis, the estimated distribution becomes more plausible and closer to the scatter plot.

In this work, we used 2-D KDE to reflect the correlation and variation tendency in the space constructed by two feature indices. As the estimation method is density based, outliers that are rather sparse would have small pixel values thus contributing little to the visual patterns. Figure 2 (b) shows a scatter plot and 2-D KDE profile of the same piece of voltage records data the afore-mentioned normal customer A. Darker areas in the estimated profile image represents locations where points emerge in high densities and outliers in scatter plot on the left side can hardly be observed in the estimated profile. Technically, the time span needed to generate distribution profile images can vary from sample to sample. If presented with more detailed event annotations as ground truth, we can generate more samples with different time durations as a kind of data augmentation. However, in our situation where specific event logs are not available, it is hard to implement such data augmentations and a fixed time duration of 10 calendar days is adopted throughout this work.

*2.2. Feature engineering*

The AC voltages and currents are three-phase data. It is fair to treat each of the three phases equivalently because here we only aim to decide whether anomalies (NTL) occurred without having to know specifically which phase is going wrong. With regard to the voltage and current, following features are extracted:

*Deviation from rated voltage*

$$\text{VD}_i = \begin{cases} \frac{V_r - V_u^i}{V_r}, & V_r \geq V_u^i \\ 0, & V_r \leq V_u^i \end{cases}, \quad (1)$$

$$\text{VD}(V_u, V_r) = \max(vd_A, vd_B, vd_C), \quad (2)$$

Where $V_r$ is the rated voltage and $V_u^i$ is the i-th phase voltage. As the voltage should remain steady for a certain customer, the maximum voltage deviation among the three phases from the above formulas can indicate the degree of how abnormal a customer is.

*Unbalance degree of voltage & current on three phases*

$$\text{AVG} = Mean(S_A, S_B, S_C) \quad (3)$$

$$\text{UD} = \frac{Mean\big(abs(S_A - \text{AVG}), abs(S_B - \text{AVG}), abs(S_C - \text{AVG})\big)}{AVG} \quad (4)$$

where, S can either be voltage or current.

Unlike the voltage, there exists no such standard value for the current. However, the discrepancy of voltage or current values on three different phases should not be too large if the system is running as expected. After all, any significant value decrease on either phases would directly result in the drop of EC.

*Active Power.* Although there are many different types of customers and their real-time active power collected every 15minutes/hour may vary greatly, we can use the contracted power to normalize it to make the power feature comparable among customers.

*Power factor.* This feature is customer-invariant and has the value range from 0 to 1.

*Load rate.* The formula is more like an approximation of the apparent powers normalized by the contracted power. And in most of the 2-dimensional distribution profile image channels, we used the load rate as the reference axis $x$.

*2.3. Generation of 'super image'*

We used a Gaussian kernel with fixed standard deviation σ and formed images as shown in figure 3. Notably, with all the features mentioned above, we can use combinations of them to get short time distribution profiles of these feature pairs. 7 grey-scale maps are generated altogether as described in

Figure 3. With each of these maps constitutes one channel, a super image with 7 channels is obtained and this is the input for the detecting neural network. Multiple samples can be generated from a customer by having a sliding window of fixed size move along the time axis of a customer's data records. The comparison between samples from statistical profile and per-row record will be shown in later section in terms of separability and final detection performance.

Before sending the super image to the network, we have to search the bounding boxes of patterns in each image channel. Position information obtained in this stage serves as auxiliary input for the network to highlight the positon of the pattern objectives and is app. We achieved this goal by setting a pixel threshold, and pixel values higher than the threshold are included in the bounding box. The threshold is obtained by experimental

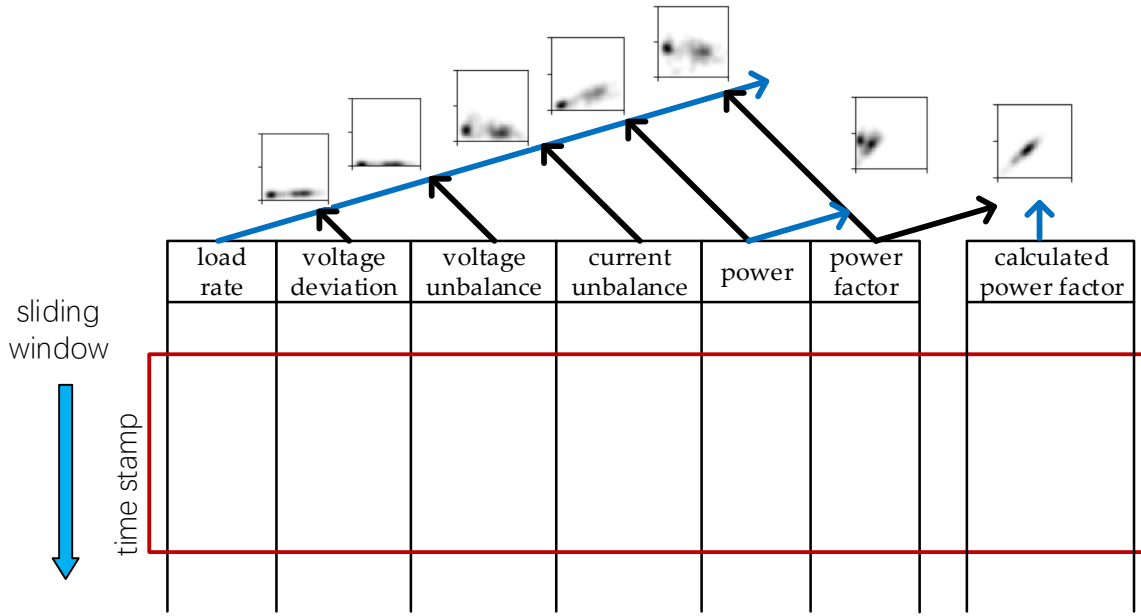

**Figure 3.** The generation of the 7-channel super image. Blue lines indicate the columns of features used as x-axis data. Here we have another feature named calculated power factor. This is the power factor we calculated with the power and load rate value.

*2.4. Framework*

The on-field inspection requires great deal of human efforts, so the number of verified customers is quite limited. Semi-supervised learning distinguishes itself by utilizing both the labeled and unlabeled data. There are different fashions of implementing the goal: self-training, co-training, transductive SVM, graph based methods and generative models. With the development deep neural networks, novel models such as generative adversarial nets [15] and tons of its variant versions emerged, which this paper will not go any deeper into. In this work, we used consistency regularization based techniques to train our model [16-19]. There are also a series of previous researches that have made significant progress in this branch of semi-supervised learning. This method does depend on any prior knowledge about the distribution of labels in class space because we have no idea about the ratio between normal and abnormal users. Throughout the paper, the ensemble strategy we adopt is from the mean teacher model [19].

The core idea of utilizing the unlabeled data here is to have a more reliable model (teacher model) to give pseudo labels to the unlabeled samples and optimize our classifier in the direction where it can give more identical predictions. The 'reliable' model we refer to in this scenario is the so called teacher model. This is achieved by taking a moving average to the parameter values of the student model. Formula (5) gives the mathematical forms of how ensembling is implemented.

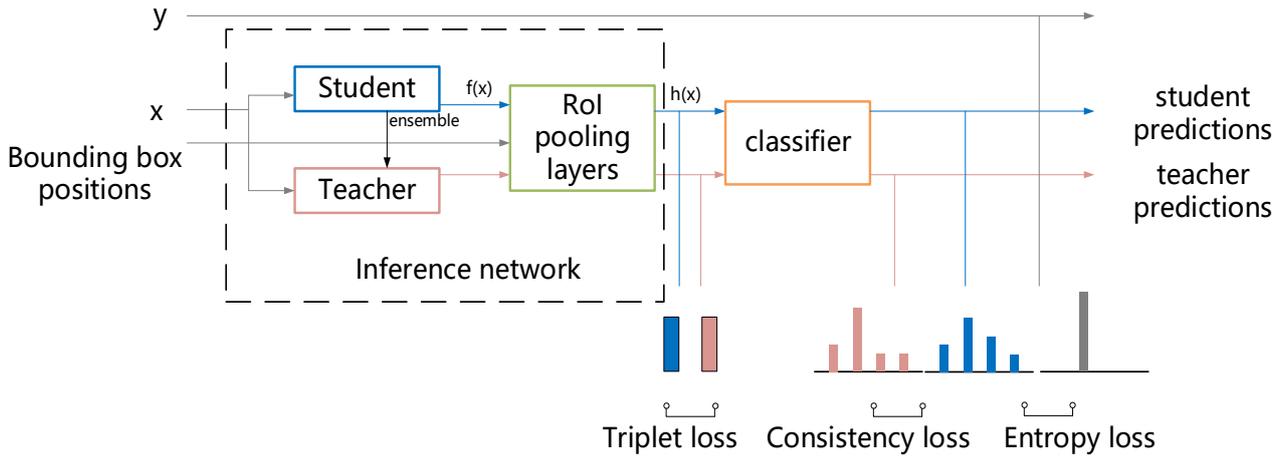

**Figure 4.** Overview of the entire model structure.

*2.4. Inference network*

We get the distribution of features to expose their visual patterns for the model to capture. For this job, convolutional based neural network has already proved its superiority and has achieved remarkable results in various image recognition and computer vision tasks [10]. Also, there exists previous work that encoded time series as images to achieve improvement encoding/decoding problems [11]. And for visual object detection problem, there are a series of famous work about two-stage object detection [12,13,14]. The region where objects lie in an image are proposed by bounding boxes for further classification. Our task shares the same purpose of highlighting the desired consumption pattern so these previous work serve as great examples of how to classify objects that occupy only part of the entire image. Figure 5 gives a detailed example of how we can decide a sample is abnormal or not. The structure of our inference network is illustrated in Figure 6.

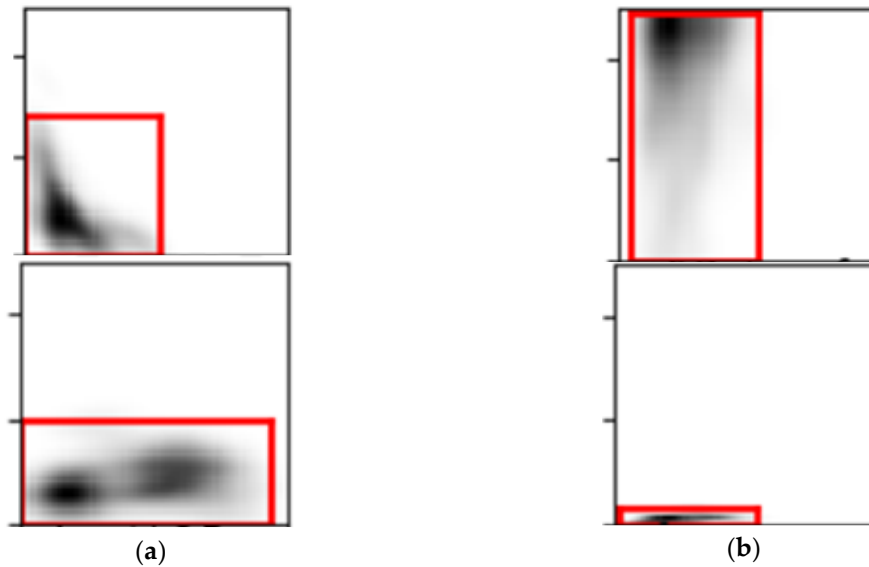

(a)　　　　　　　　　　　　　(b)

**Figure 5.** (a) The distributions of the 2-D points (load rate, current unbalance degree) from a normal (upper) and an abnormal (lower) customer. (b) The distributions of 2-D points (load rate, power factor) and (load rate, power) from a same abnormal user. The ranges are marked with red bounding boxes. Normally, the unbalance degree decreases as the load rate grows as shown in the left in (a). Because as the current on three phases grows, small disturbance would not have much effect on it. So the unbalance degree would turn low and steady. And the right figure contradicts to this fact. In (b), the power factor is 1 for the most of the time, but the power data is nearly zero. This is a typical sign of customers' conducting theft.

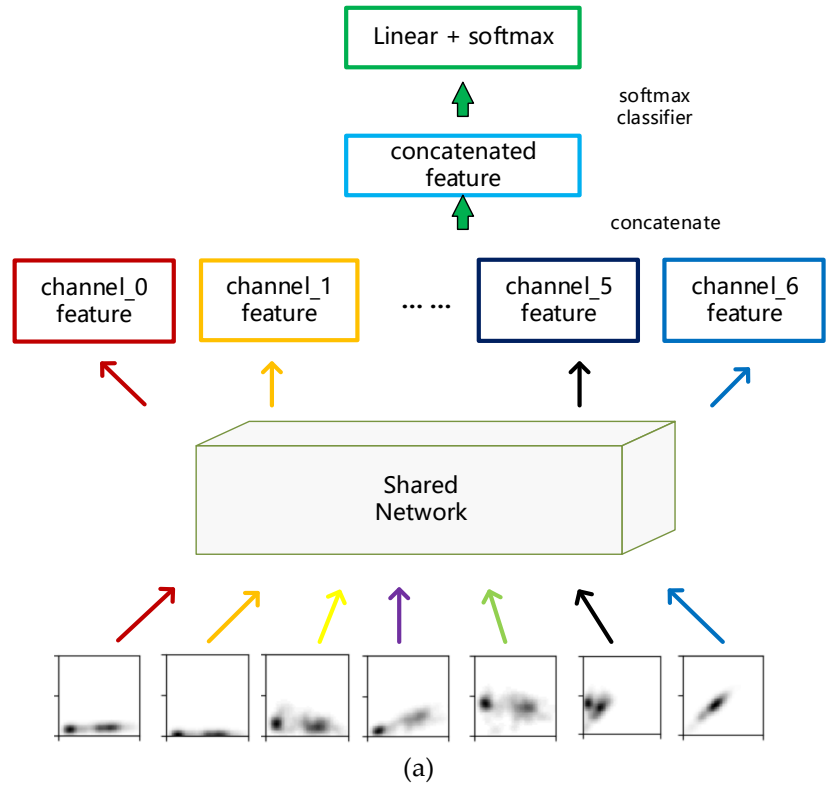

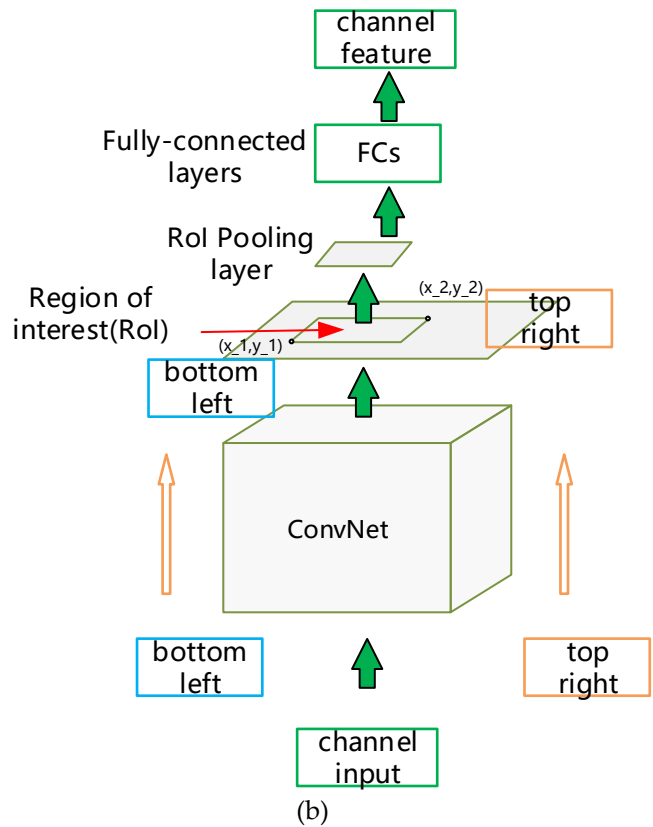

**Figure 6.** (a)The structure of the inference network. (b) The architecture of the shared network.

Figure 6 (a) is the overall structure of our inference network. Although each statistical profile is treated as a channel of a super image, the information each of these channels carries, such as pattern shapes, the regions patterns lie in, are totally different and should be processed separately .Every channel in a super image goes through a shared network whose internal structure is shown in figure 6

(b). The internal structure of the shared network is much like that in [13] except that we search the region of interests here by setting a pixel value threshold to get a rectangular area where the pattern lies. So there is only one bounding box for an input channel. After obtaining every channel feature, we concatenate them together to get a final feature representation of an input super image. The convolutional based network denoted as 'ConvNet' used the same set of parameters to extract graphic information of all the image channels and its detailed information is listed in Table 2.

Table 2. The architecture of the convolutional based layers

| |
|---|
| Input 50×50×1 (one channel of a super image) |
| Gaussian noise σ = 0.15 |
| 3×3 conv. 32 lReLU($\alpha = 0.1$) same padding followed by BN |
| 3×3 conv. 32 lReLU($\alpha = 0.1$) same padding followed by BN |
| 2×2 max-pool, dropout 0.5 |
| 3×3 conv. 64 lReLU($\alpha = 0.1$) same padding followed by BN |
| 3×3 conv. 64 lReLU($\alpha = 0.1$) same padding followed by BN |
| 2×2 max-pool, dropout 0.5 |
| 3×3 conv. 128 lReLU($\alpha = 0.1$) same padding followed by BN |
| 3×3 conv. 128 lReLU($\alpha = 0.1$) same padding followed by BN |
| 2×2 max-pool, dropout 0.5 |
| 3×3 conv. 256 lReLU($\alpha = 0.1$) same padding followed by BN |
| 1×1 conv. 128 lReLU($\alpha = 0.1$) same padding followed by BN |
| 1×1 conv. 64 lReLU($\alpha = 0.1$) same padding followed by BN |

*2.5. Ensembling method and consistency loss*

Formally, let θ denote the weight parameters in the neural network and $\theta_t$ denote the parameters at training step t. And the ensembling strategy is as follow:

$$\theta'_t = \alpha \theta'_{t-1} + (1 - \alpha)\theta_t, \tag{5}$$

Then, consistency cost is introduced to minimize the prediction difference between the teacher and student model:

$$J(\theta) = \mathbb{E}_{x,\eta',\eta}[\|f(x, \theta', \eta') - f(x, \theta, \eta)\|], \tag{6}$$

Where, $\eta$ is the noise applied to input images.

Apart from the consistency cost, we introduced another loss item, triplet loss [22]. This is originally proposed for face recognition tasks. However the goal of face verification is more demanding than general classification problems and what triplet loss dose is to constrain points of same classes to have consistent embeddings while push points of different classes far away in the embedding space. (6) is an explanatory formula of how triplet loss works:

$$l_G(i,j) = \begin{cases} \|h(x_i) - h(x_j)\|^2 & if\ y_i = y_j \\ \max(0, margin - \|h(x_i - h(x_j)\|^2) & if\ y_i = y_j \end{cases}, \tag{7}$$

Additionally, the labels in the formula are not all ground truth in labeled data set and the triplet loss is applied for all the input in a training batch. For those input without labels, we use the predictions given by the teacher model as pseudo labels.

*2.6 training algorithm*

With the frameowork and the loss items introduced, the pseudo code of our training algorithm is presented in Algorithm 1. Notably, the coordinates of bottom left and top right vertices of bounding boxes are sent to the network as auxiliary input. The RoI pooling operation is executed on the feature maps and the coordinates of bounding boxes are linearly projected on the feature map. As for our strategy of drawing mini-batches, we separately draw samples from labeled data set and unlabeled data set by random while maintaining the ratio of 1:3 between labeled an unlabeled samples.

---

**Algorithm 1** Training algorithm

---
**Require**: $x_i$ = input images
**Require**: L = set of training input indices with known labels
**Require**: $y_i$ = labels for labeled inputs, i ∈ L
**Require**: $bb_i$ = bounding boxes, i ∈ B
**Require**: W_u= unsupervised loss weight
**Require**: $f_\theta(x)$ = neural network with trainable parameters $\theta$ as student model
**Require**: $f_{\theta'}(x)$ = neural network as teacher model whose parameter $\theta'$ with initial value $\theta$
**Require**: α = moving average momentum for parameters
**Require**: $\eta$ = random Gaussian noise added to the input
**for** t in [1, num_iterations] **do**
   draw a mini-batch B from labeled and unlabeled samples randomly
   $f_i \leftarrow f_\theta(x_{i \in B}, bb_{i \in B}, \eta)$ evaluate network outputs
   $f'_i \leftarrow f_{\theta'}(x_{i \in B}, bb_{i \in B}, \eta')$ evaluate network outputs
   Find triplets T of components <i, j, k> in B where $y_i = y_j$ and $y_i \neq y_k$; $y_i$ is pseudo label given by    $f'_i$ if $i \notin (B \cap L)$
   loss ← $-\frac{1}{|B|}\sum_{i\in(B\cap L)} \log f_i[y_i]$
        + W_u$\frac{1}{C|B|}\sum_{i\notin(B\cap L)}\|f_i - f'_i\|$ + W_u$\frac{1}{|T|}(\sum_{i,j,k\in B} l_G(i,j) + l_G(i,k))$
   update $\theta$ using ADAM optimizer
   update $\theta'$ by $\theta'_t = \alpha\theta'_{t-1} + (1-\alpha)\theta_t$
**end for**
return $\theta$

---

## 3. Results

*3.1. Experiment settings and metrics*

    The data used in our experiment is from real-world SM records. We use the data from a power company in China to train and validate our model. There are 193 verified and 2929 unverified customers in total, among which 54 of them are labeled as abnormal customers by on-field inspection whatever the causes are and 139 of them are labeled as normal ones.. As is introduced in section 2, data records of a customer can be transformed into numbers of samples as the sliding window moves along the time axis. Here we adopt the sliding window of 10-day time range regardless of the features' sampling frequency and the overlap between windows is 5 days, that is to say, each image channel of a sample consists of 240 points if the corresponding feature is collected hourly. As a result, it would make 8797 labeled training samples and more than 130000 unlabeled samples in total. However, in order to significantly cut down time for training, we randomly selected 50000 samples from part of the unlabeled users. In order to justify the generalization ability of our method, we split the verified customers randomly while maintaining the ratio between the number of samples in training and validation set, which is approximately 1:3. And this process is repeated for three times.

Table 3. Description of data set

| Customer label | Number of customers | Number of data records | Number of samples |
|---|---|---|---|
| Normal | 139 | 1145832 | 7685 |
| NTL detected | 54 | 137112 | 1112 |
| Unlabeled | 2929 | 16390248 | 132481 |

For our detection problem where the negative samples are far more than positive ones, overall precision is not a suitable criterion to assess the performance of our algorithm. However, the precision and recall of the positive samples alone would give a direct impression of the detection results:

$$Presion = \frac{TP}{TP+FP}, \tag{8}$$

$$Recall = \frac{TP}{TP+FN}, \tag{9}$$

$$F1 = \frac{2 \times Presion \times Recall}{Precision+Recall}, \tag{10}$$

where TP(True positive), FP(False positive), FN(False negative) are the number of NTL samples that are successfully detected, the number of normal samples that are classified as NTL and the number of NTL samples that are classified as normal ones. F1 score is an overall judgement combining the precision and recall.

As is in most binary classification problems where the output is a two-dimensional vector that gives the corresponding probability of each class the input belongs to, we take the hypothesis with higher probability as the prediction. In other words, the decision threshold is 0.5. If we are to observe the overall performance of our algorithm by varying the threshold, True Positive Rate (TPR) and False Positive Rate (FPR) are needed to draw the ROC curve as well as get the area under the ROC curve, AUC score [21]:

$$TPR = \frac{TP}{TP+FN}, \tag{11}$$

$$FPR = \frac{FP}{TN+FP}, \tag{12}$$

*3.2. Separability*

To justify our method of transforming the data, we visualized separability of the training data using a visualization technique called t-SNE [20]. It non-linearly projects features from high-dimensional space to lower-dimensional space while trying to maintain their original relative positions. As shown in Figure 7, (a) is the 2-d projection of the 6-D feature vector (load rate, voltage deviation, voltage unbalance degree, current unbalance degree, power factor and power) and each point represents the features at one sampling time stamp in validation set. (b) uses the same features but takes longer time range of 10 days to form each point in the figure. Points in (c) are our proposed 7-channel 'super images' generating from 10-day time range of features..

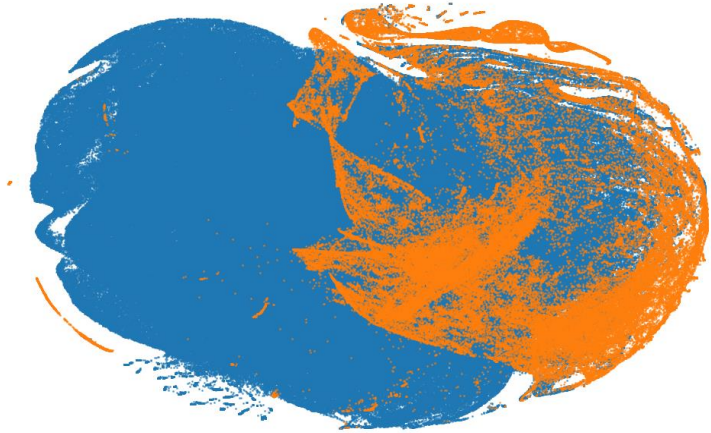

(**a**)

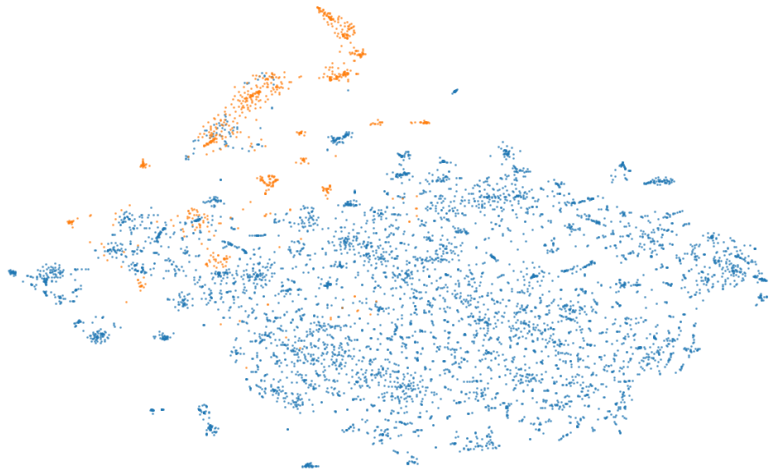

(**b**)

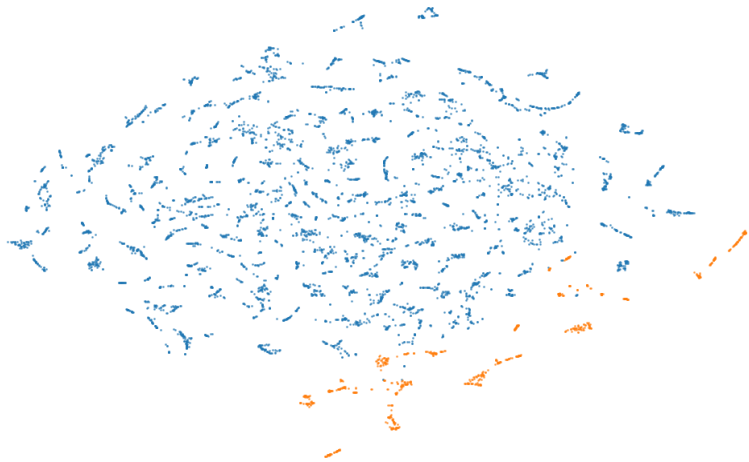

(**c**)

**Figure 7.** t-SNE visualization comparison of different ways of generating data samples. Points in orange and blue indicate abnormal and normal types respectively. Notably, the number of points in (a) is much more that those in (b) and (c). This is because each sample in (b) contains a certain time range of raw features in (a).

Obviously, our transformation method makes the classification problem more feasible as the result in Figure 7 (a) shows that it is nearly impossible to separate samples of different labels. This is a proof of what was discussed earlier that viewing the data in the longer term is a better solution to address the detection problem.

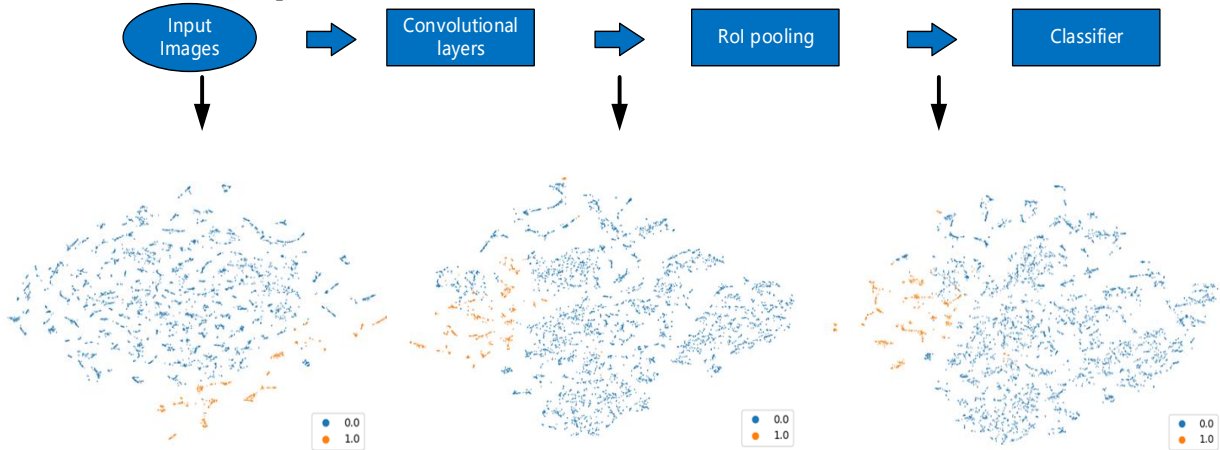

**Figure 8.** Separability of samples in validation set changes as samples go through the inference network

From Figure 8 we can see that as samples go through the inference network, points of the sample labels gradually gather together, especially for the abnormal samples which are plotted in orange. This is another evidence that our data transformation, together with such network structure, works pretty fine for our classification goal.

*3.2. Detection results*

Verified customers are randomly split into training set and validation set for three separate times and the training and validation process is carried out based on the data. Detection accuracy is not suitable to assess our algorithm performance because in this situation where negative samples are in absolute majority, the overall accuracy would still seem rather high even if our algorithm fails to find out any positive samples. However, Receiver Operation Characteristic (ROC) and AUC [21] score would be a better choice. It describes the relationship between true positive rate and false positive rate as the decision thresholds vary.

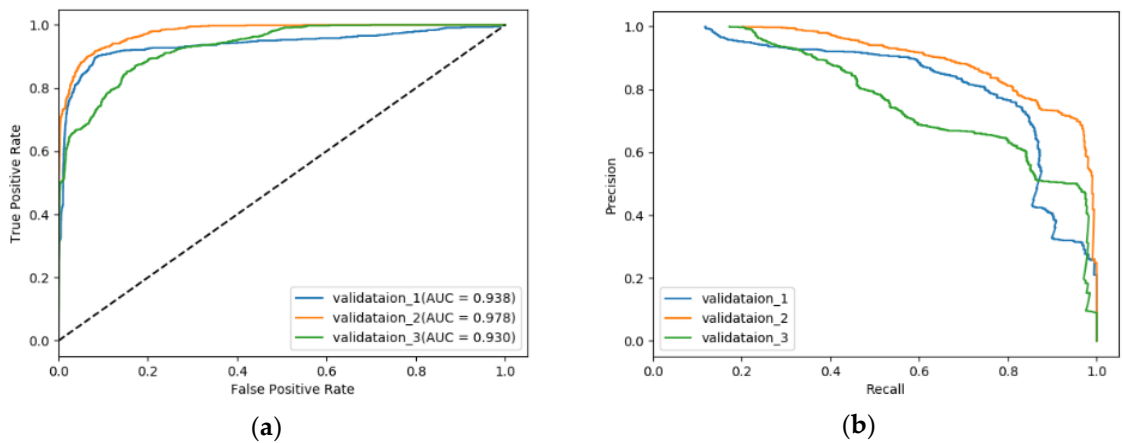

**Figure 9.** (a) ROC curves (b) Precision-Recall curves in 3 experiment rounds on different training/validation sets.

Table 4. Precision and recall of positive samples

| Experiment# | Precision | Recall | F1-score | Overall AUC score | Positive/Total |
|---|---|---|---|---|---|
| Validation_1 | 0.84 | 0.74 | 0.79 | 0.9379 | |
| Validation_2 | 0.97 | 0.88 | 0.92 | 0.9775 | 785/6691 |
| Validation_3 | 0.84 | 0.80 | 0.82 | 0.9304 | |

Similar to any other anomaly detection scenarios, our data set is quite imbalanced where most of the data is negative/normal in both training and validation set. So apart from overall performance, we are particularly interested in our algorithm's performance on positive/abnormal samples. The precision and recall results of positive samples in different experiment rounds are listed in Table 4.

For our NTL detection problem, the characteristics of data records from different customers can be very different because there a variety of types of customers. This fact can be directly reflected in the detection results shown in Figure 9. For different train/validation splits, differences in terms of ROC curves, AUC score or Precision-Recall curves are all non-negligible. This is totally understandable and we suppose it is due to our random selection strategy: sometimes the customers chosen for training can well cover most of the abnormal causes of NTL while sometimes the customers for training have only limited pattern of manifestation. However, all the detection results are pretty good from multiple perspectives thus we believe our method has its specialties.

Table 5. Comparison of semi-supervised and supervised learning method

| Model | Label | Precision | Recall | F1-score |
|---|---|---|---|---|
| Conv9 + semi-supervised | Normal | 0.97 | 0.98 | 0.96 |
| | NTL | 0.84 | 0.74 | 0.79 |
| Conv9 + supervised | Normal | 0.88 | 1.00 | 0.94 |
| | NTL | 0.00 | 0.00 | 0.00 |
| Resnet21 + supervised | Normal | 0.89 | 1.00 | 0.94 |
| | NTL | 1.00 | 0.03 | 0.06 |

Our forward inference network consists of 9 convolutional layers, which is denoted Conv9 in Table 5. We justify that utilizing the unlabeled samples is essential by comparing the results of models trained with or without semi-supervised learning strategy in Table 5. Even though we used some sampling strategy to make sure that in a training batch, the numbers of positive and negative samples are throughout the experiments, results shows that sole labeled training data is not enough for a model to perform well on a larger and maybe more complicated validation samples. To guarantee that our inference network is not the cause of the poor performance of supervised model, we used a more traditional and reliable architecture, resnet21. It turns out it still cannot address the problem that a supervised model basically classify every sample as negative.

Semi-supervised learning methods help find a more reasonable decision boundary with the presence of unlabeled data thus preventing the overfitting problem caused by the limited labeled data. Figure 10 gives the results when we change the size of labeled data in training set while keeping other settings unchanged. Generally speaking, the anomaly detection performance gradually increase as the size of labeled samples grows from 500 to 1500. There is a slight drop in recall rate when the size of labeled data changes from 500 to 1000, but the precision increased a bit.

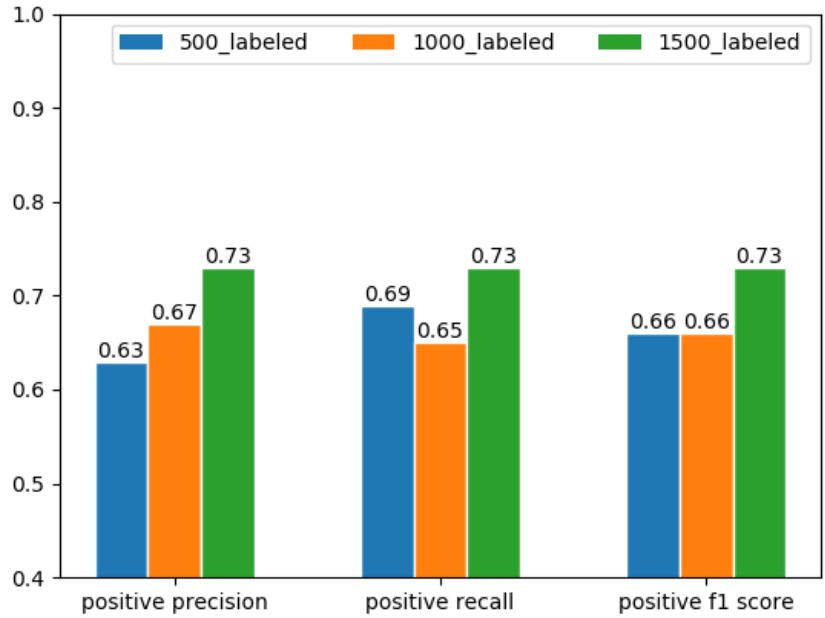

**Figure 10.** The precision, recall and f1 score of positive samples on the same validation set as the number of labeled samples changes. The results are obtained by randomly selecting part of total labeled training samples and use the same model and semi-supervised training method.

In order to prove that our network architecture is different and more suitable for the detection problem and especially for our designated data format, we did a comparison with the original mean teacher model by gradually adding our adaptions.

**Table 6.** Results by different model settings

| Model | Label | Precision | Recall | F1-score |
|---|---|---|---|---|
| Mean teacher | Normal | 0.91 | 1.00 | 0.95 |
| | NTL | 0.88 | 0.23 | 036 |
| Mean teacher + Triplet loss | Normal | 0.94 | 0.98 | 0.96 |
| | NTL | 0.74 | 0.49 | 0.59 |
| Mean teacher+Triplet loss+RoI pooling | Normal | 0.97 | 0.98 | 0.96 |
| | NTL | 0.87 | 0.74 | 0.79 |

As we can see from Table 6, simpler models tend to categorize samples into the class that is in the majority. For the general NTL detection goal, it is equivalently important have a high value on both precision and recall. Extreme result on either single index is obviously not desired. In that sense, our network architecture does improve the performance of detection. An intuitive explanation of such improvement is that: triplet loss tries to minimize the pattern discrepancies among users in a same category while the role RoI pooling plays is that it highlights where the points cloud is in an image channel and naturally makes the pattern features more explicit.

*3.3.Discussion*

The major challenges for most of the NTL detection problems are the anomalies can take on various forms of manifestations that are hard to capture. Worse still, the number of cases we can study from is often limited as result of the labor-consuming on-field inspection. So it not a traditional anomaly detection problem and it is essential to make use of all the types of data available to get complete while prominent representations of the input. Most of the previous work mainly focus on the historical energy consumption data because it has the most direct relevance to NTL detection problem. However, the types of NTL that can be detected are bound to be limited by the limited choice of measurements. [2,3,4,5,6] used sole energy consumption (EC) data. Apart from it, [23] utilized auxiliary databases and [13] made use of credit worthiness ratings. [25] tried to gather different features form various types of SM data such as the quality of measurements, electrical magnitudes, GIS data and technological characteristics of the SM besides EC. As for our proposed method, we utilize as many types of data as we have access to and mainly focus on the electrical parameters. And we considered the statistical characteristics of the related features, which are the objects our model tries to classify.

Our proposed approach deals with the situation where there are only a small number of labeled samples. There are different fashion of detecting NTL among these work, for example, classification based [6,7,23,25], statistical analysis based [4,5], clustering based[24] or merely judgements with threshold values[3]. Most of these work took the advantage of the abundance of customer type labels. [24] compared the clustering results of similar customers. And for the classification based methods, they treated the problem as a totally supervised one. Figure 10 shows that although the detection performance may suffer with the decrease of the number of labeled samples, acceptable results can still be obtained.

Another advantage of our method is that the time range we need to generate a sample is rather short compared with other statistics based method. This allows us to make quicker responses to the emergent NTL attacks. Besides, our data transformation approach can tolerate slight data distortion or missing problem and we can generate more available samples.

**Table 7.** Comparison with the state-of-the-art

| Criteria | [23] | [13] | [24] | [9] | [25] | Ours |
|---|---|---|---|---|---|---|
| Data privacy | low | high | high | medium | medium | high |
| Data types | Monthly EC | Monthly EC & auxiliary databases | Monthly EC & auxiliary databases | Monthly EC & auxiliary databases | SM data & auxiliary databases | Hourly SM data |
| Detection delay | 12 months | 12 months | 12 months | - | 90 days | 10 days |
| AUC score | 0.56 | 0.63 | 0.74 | 0.84 | 0.91 | 0.94 |

Considering the different data sets and different proportions of positive/negative samples, it neither fair nor suitable to do comparisons based on quantitative criteria such as AUC score and precision/recall. But as for the data privacy, the transformation we did hides the original information that SM collect. And for a trained model, we can use a certain range of data records to form our data sample to sufficiently cut down the detection delay.

There are still some weaknesses or works we have not done sufficiently for now. Even though we aim to cover as many types of anomalies as possible in the detecting stage, we have not come up with a solution to explicitly indicate the exact causes of NTLs. And our detection results strongly depend on the coverage of different forms of NTL attacks used in the labeled training set.

## 5. Conclusions

This paper presented a thorough methodology for detecting non-technical losses. The transformation from the data records collected by smart meters into the super images allows us to view the consumption behaviors of a customer from a statistical perspective in the longer term. The new data format also provides a different way of extracting features and integrated analysis of more types of features would have greater potential of detecting wider range of anomalies. Followed by a priori knowledge free semi-supervised learning strategy, our method demonstrated its superiority to the supervised learning in the situation where labeled data is in the minority of the entire data set.

Our method is trained and validated on the realistic data from a power grid of China. With reference to some ideas in two stage object detection models, we designed our network architecture to effectively capture features for classification. Ablation studies in the experiment section demonstrates that our method in each stage dose work out and the comparison with the state-of-the art methodologies proves that our result is rather competitive and our method has its advantages.